\documentclass[11pt]{article}
\usepackage{authblk}
\usepackage{eacl2017}
\usepackage{times}
\usepackage{latexsym}
\usepackage{multirow}
\usepackage{hyperref}
\usepackage{relsize}
\usepackage{url}
\usepackage{amssymb}
\usepackage{booktabs}

\eaclfinalcopy
\def\eaclpaperid{226}

\usepackage[textsize=tiny]{todonotes}

\newcommand{\real}{\mathbb{R}}

\title{
    Neural Architectures for Fine-grained Entity Type Classification
}

\author{
    Sonse Shimaoka$^{\dagger}$\thanks{
        \hspace{0.55em}This work was conducted during a research visit
        to University College London.
    } \hspace{0.75em}
    Pontus Stenetorp$^{\ddagger}$ \hspace{0.75em}
    Kentaro Inui$^{\dagger}$ \hspace{0.75em}
    Sebastian Riedel$^{\ddagger}$ \\
    {\tt \{simaokasonse,inui\}@ecei.tohoku.ac.jp} \\
    {\tt \{p.stenetorp,s.riedel\}@cs.ucl.ac.uk} \\
    $^{\dagger}$Graduate School of Information Sciences, Tohoku University \\
    $^{\ddagger}$Department of Computer Science, University College London
}

\date{}

\begin{document}
\maketitle
\begin{abstract}
    In this work, we investigate several neural network architectures for fine-grained entity type classification and make three key contributions.
    Despite being a natural comparison and addition, previous work on attentive neural architectures have not considered hand-crafted features and we combine these with learnt features and establish that they complement each other.
    Additionally, through quantitative analysis we establish that the attention mechanism learns to attend over syntactic heads and the phrase containing the mention, both of which are known to be strong hand-crafted features for our task.
    We introduce parameter sharing between labels through a hierarchical encoding method, that in low-dimensional projections show clear clusters for each type hierarchy.
    Lastly, despite using the same evaluation dataset, the literature frequently compare models trained using different data.
    We demonstrate that the choice of training data has a drastic impact on performance, which decreases by as much as 9.85\% loose micro F1 score for a previously proposed method.
    Despite this discrepancy, our best model achieves state-of-the-art results with 75.36\% loose micro F1 score on the well-established \textsc{Figer (GOLD)} dataset and we report the best results for models trained using publicly available data for the OntoNotes dataset with 64.93\% loose micro F1 score.
\end{abstract}

\section{Introduction}

\begin{figure}[ht]
    \includegraphics[width=8cm]{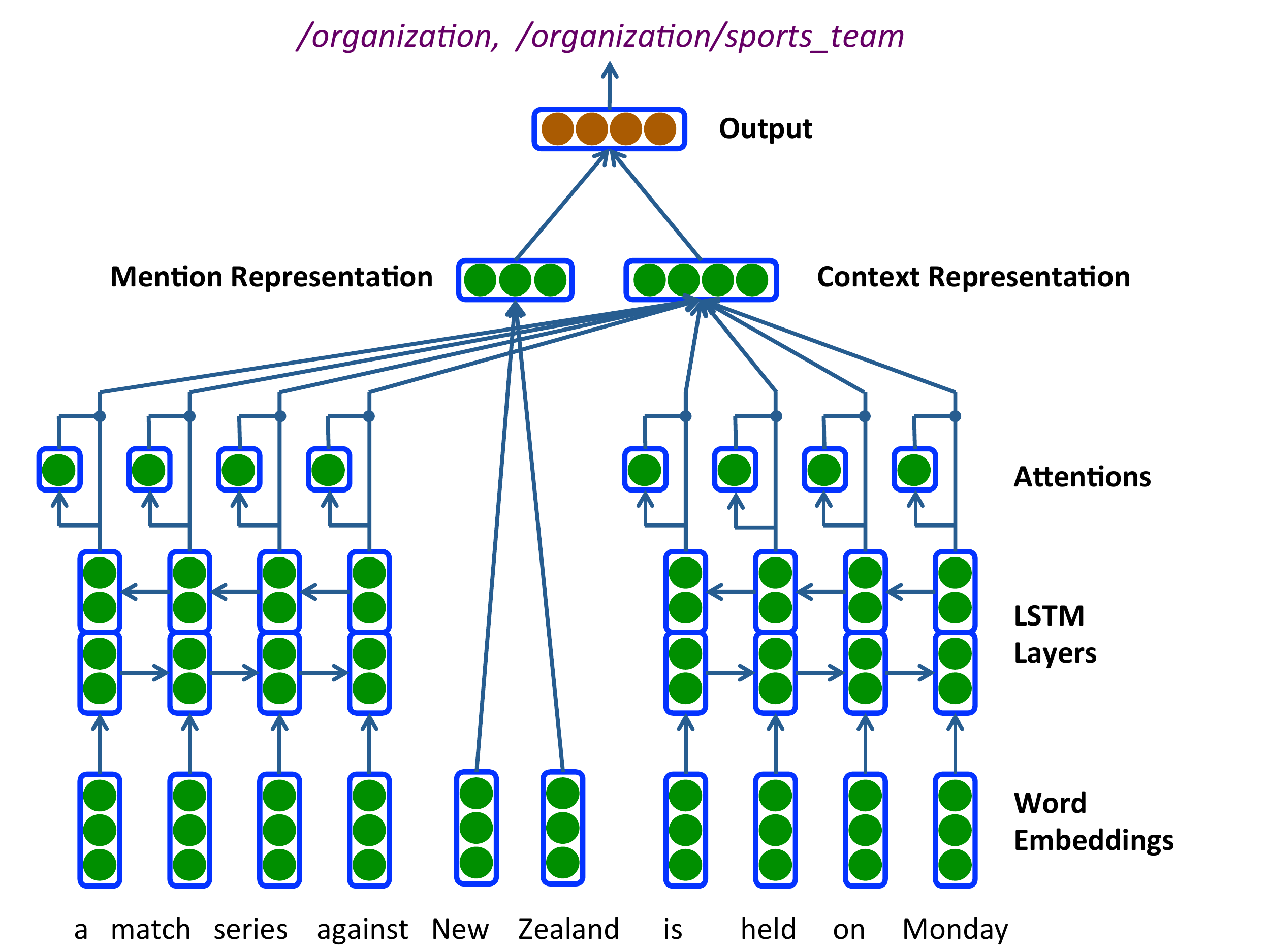}
    \centering
    \caption{
        An illustration of the attentive encoder neural model predicting
        fine-grained semantic types for the mention ``New Zealand'' in the
        expression ``a match series against New Zealand is held on Monday''.
    }
    \label{fig:model}
\end{figure}

Entity type classification aims to label entity mentions in their context with their respective semantic types.
Information regarding entity type mentions has proven to be valuable for several natural language processing tasks; such as question answering~\cite{lee2006fine}, knowledge base population~\cite{carlson2010coupled}, and  co-reference resolution~\cite{recasens2013life}.
A natural extension to traditional entity type classification has been to divide the set of types -- which may be too \textit{coarse-grained} for some applications~\cite{sekine2008extended} -- into a larger set of \textit{fine-grained} entity types~\cite{lee2006fine,ling2012fine,yosef2012hyena,gillick2014context,del2015finet}; for example \texttt{person} into \texttt{actor}, \texttt{artist}, etc.

Given the recent successes of attentive neural models for information extraction~\cite{globerson2016collective,shimaoka2016attentive,yang2016hierarchical}, we investigate several variants of an attentive neural model for the task of fine-grained entity classification (e.g. Figure~\ref{fig:model}).
This model category uses a neural attention mechanism -- which can be likened to a soft alignment -- that enables the model to focus on informative words and phrases.
We build upon this line of research and our contributions are three-fold:
\begin{enumerate}
    \item{
    Despite being a natural comparison and addition, previous work on attentive neural architectures do not consider hand-crafted features.
    We combine learnt and hand-crafted features and observe that they complement each other.
    Additionally, we perform extensive analysis of the attention mechanism of our model and establish that the attention mechanism learns to attend over syntactic heads and the tokens prior to and after a mention, both which are known to be highly relevant to successfully classifying a mention.
    }
    \item{
    We introduce label parameter sharing using a hierarchical encoding that improves performance on one of our datasets and the low-dimensional projections of the embedded labels form clear coherent clusters.
    }
    \item{
    While research on fine-grained entity type classification has settled on using two evaluation datasets, a wide variety of training datasets have been used -- the impact of which has not been established.
    We demonstrate that the choice of training data has a drastic impact on performance, observing performance decreases by as much as 9.85\% loose Micro F1 score for a previously proposed method.
    However, even when comparing to models trained using different datasets we report state-of-the-art results of 75.36\% loose micro F1 score on the \textsc{Figer (GOLD)} dataset.
    } 
\end{enumerate}

\section{Related Work}

Our work primarily draws upon two strains of research, fine-grained entity
classification and attention mechanisms for neural models.
In this section we introduce both of these research directions.

By expanding a set of coarse-grained types into a set of 147 fine-grained types,
\newcite{lee2006fine} were the first to address the task of fine-grained entity
classification.
Their end goal was to use the resulting types in a question answering system and
they developed a conditional random field model that they trained and evaluated on a
manually annotated Korean dataset to detect and classify entity mentions.
Other early work include \newcite{sekine2008extended}, that emphasised the need
for having access to a large set of entity types for several NLP applications.
The work primarily discussed design issues for fine-grained set of entity
types and served as a basis for much of the future work on fine-grained entity
classification.

The first work to use
distant supervision \cite{mintz2009distant} to induce a large -- but noisy --
training set and manually label a significantly smaller dataset to evaluate
their fine-grained entity classification system, was \newcite{ling2012fine} who
introduced both a training and evaluation dataset \textsc{Figer (GOLD)}.
Arguing that fine-grained sets of types must be organised in a very fine-grained
hierarchical taxonomy, \newcite{yosef2012hyena} introduced such a taxonomy
covering $505$ distinct types.
This new set of types lead to improvements on \textsc{Figer (GOLD)},
and they also demonstrated that the fine-grained labels could be used as
features to improve coarse-grained entity type classification performance.
More recently, continuing this very fine-grained strategy,
\newcite{del2015finet} introduced the most fine-grained entity type
classification system to date, covering the more than $16,000$ types contained
in the WordNet hierarchy.

While initial work largely assumed that mention assignments could be done
independently of the mention context, \newcite{gillick2014context} introduced the
concept of context-dependent fine-grained entity type classification where the
types of a mention is constrained to what can be deduced from its context and
introduced a new OntoNotes-derived manually annotated evaluation dataset.
In addition, they addressed the problem of label noise induced by distant
supervision and proposed three label cleaning heuristics.
Building upon the noise reduction aspects of this work, \newcite{ren2016label}
introduced a method to reduce label noise even further, leading to significant
performance gains on both the evaluation dataset of \newcite{ling2012fine} and
\newcite{gillick2014context}.

\newcite{yogatama2015embedding} proposed to map hand-crafted features and labels
to embeddings in order to facilitate information sharing between both related
types and features.
A pure feature learning approach was proposed by \newcite{dong2015hybrid}.
They defined $22$ types and used a two-part neural classifier that used a
recurrent neural network to obtain a vector representation of each entity
mention and in its second part used a fixed-size window to capture the context
of a mention.
A recent workshop paper \cite{shimaoka2016attentive} introduced an attentive neural
model that unlike previous work obtained vector representations for each mention
context by composing it using a recurrent neural network and employed an
attention mechanism to allow the model to focus on relevant expressions in the
mention context.
Although not pointed in \newcite{shimaoka2016attentive}, the attention mechanism used differs from previous work in that it does not condition the attention.
Rather, they used  global weights optimised to provide attention for every fine-grained entity type classification decision.

To the best of our knowledge, the first work that utilised an attention architecture within the context of NLP was \newcite{bahdanau2014neural}, that allowed a machine translation decoder to attend over the source sentence.
Doing so, they showed that adding the attention mechanism significantly improved their machine translation results as the model was capable of learning to align the source and target sentences.
Moreover, in their qualitative analysis, they concluded that the model can correctly align mutually related words and phrases. 
For the set of neural models proposed by \newcite{hermann2015teaching}, attention mechanisms are used to focus on the aspects of a document that help the model answer a question, as well as providing a way to qualitatively analyse the inference process.
\newcite{rocktaschel2015reasoning} demonstrated that by applying an attention mechanism to a textual entailment model, they could attain state-of-the-art results, as well as analyse how the entailing sentence would align to the entailed sentence.

Our work differs from previous work on fine-grained entity classification in that we use the \textit{same publicly available training data} when comparing models.
We also believe that we are the first to consider the \textit{direct combination of hand-crafted features and an attentive neural model}.

\section{Models}

In this section we describe the neural model variants used in this paper as well as a strong feature-based baseline from the literature.
We pose fine-grained entity classification as a multi-class, multi-label classification problem.
Given a mention in a sentence, the classifier predicts the types $t \in \{1,0\}^K$ where $K$ is the size of the set of types.
Across all the models, we compute a probability $y_k \in \real$ for each of the $K$ types using logistic regression.
Variations of the models stem from the ways of computing the input to the logistic regression.

At inference time, we enforce the assumption that at least one type is assigned to each mention by first assigning the type with the largest probability.
We then assign any additional types based on the condition that their corresponding probabilities must be greater than a threshold of $0.5$, which was determined by tuning it using development data.

\subsection{Sparse Feature Model}

\begin{table}
\centering
\smaller[2]
\begin{tabular}{lll}
    Feature     & Description                               & Example           \\
    \toprule
    Head        & Syntactic head of the mention             & Obama             \\
    Non-head    & Non-head words of the mention             & Barack, H.        \\
    Cluster     & Brown cluster for the head token          & 1110, \ldots      \\
    Characters  & Character trigrams for the mention head   & :ob, oba, \ldots  \\
    Shape       & Word shape of the mention phrase          & Aa A. Aa          \\
    Role        & Dependency label on the mention head      & subj              \\
    Context     & Words before and after the mention        & B:who, A:first    \\
    Parent      & The head's lexical parent                 & picked            \\
    Topic       & The LDA topic of the document             & LDA:13            \\
\end{tabular}
\caption{
    Hand-crafted features, based on those of Gillick~et~al.~(2014), used 
    by the sparse feature and hybrid model variants in our experiments.
    The features are extracted for each entity mention and the example mention
    used to extract the example features in this table is ``\ldots who [Barack
    H. Obama] first picked \ldots''.
}
\label{tbl:features}
\end{table}

For each entity mention $m$, we create a binary feature indicator vector $f(m) \in \{0,1\}^{D_f}$ and feed it to the logistic regression layer.
The features used are described in Table~\ref{tbl:features}, which are comparable to those used by \newcite{gillick2014context} and \newcite{yogatama2015embedding}.
It is worth noting that we aimed for this model to resemble the independent classifier model in \newcite{gillick2014context} so that it constitutes as a meaningful well-established baseline; however, there are two noteworthy differences.
Firstly, we use the more commonly used clustering method of \newcite{brown1992class}, as opposed to \newcite{uszkoreit2008distributed}, as \newcite{gillick2014context} did not make the data used for their clusters publicly available.
Secondly, we learned a set of 15 topics from the OntoNotes dataset using the LDA \cite{blei2003latent} implementation from the popular gensim software package,\footnote{\url{http://radimrehurek.com/gensim/}} in contrast to \newcite{gillick2014context} that used a supervised topic model trained using an unspecified dataset.
Despite these differences, we argue that our set of features is comparable and enables a fair comparison given that the original implementation and some of the data used is not publicly available.

\subsection{Neural Models}

The neural models from \newcite{shimaoka2016attentive} processes embeddings of the words of the mention and its context; and we adopt the same formalism when introducing these models and our variants.
First, the mention representation $v_m \in \real^{D_m \times 1}$ and context representation $v_c \in \real^{D_c \times 1}$ are computed separately. 
Then, the concatenation of these representations is used to compute the prediction:

\begin{equation}
    y = \frac{1}{1 + \exp \left( -W_y \left[ \begin{array}{c} v_{m} \\ v_{c}\\ \end{array} \right] \right)}
\end{equation}

Where $W_y \in \real^{K \times (D_m + D_c )}$ is the weight matrix.

Let the words in the mention be $m_1, m_2,..., m_{|m|}$.
Then the representation of the mention is computed as follows:

\begin{equation}
    v_m = \frac{1}{|m|}\sum_{i=1}^{|m|} u(m_i)
\end{equation}

Where $u$ is a mapping from a word to an embedding.
This relatively simple method for composing the mention representation is motivated by it being less prone to overfitting.

Next, we describe the three methods from \newcite{shimaoka2016attentive} for computing the context representations; namely, Averaging, LSTM, and Attentive Encoder.

\subsubsection{Averaging Encoder}

Similarly to the method of computing the mention representation, the Averaging encoder computes the averages of the words in the left and right context.
Formally, let $l_{1},...,l_{C}$ and $r_1,...,r_C$ be the words in the left and right contexts respectively, where $C$ is the window size.
Then, for each sequence of words, we compute the average of the corresponding word embeddings.
Those two vectors are then concatenated to form the representation of the context $v_c$.


\subsubsection{LSTM Encoder}

For the LSTM Encoder, the left and right contexts are encoded by an LSTM \cite{hochreiter1997long}.
The high-level formulation of an LSTM can be written as:

\begin{equation}
    h_i, s_i = lstm(u_i,h_{i-1},s_{i-1})
\end{equation}

Where $u_i \in \real^{D_m \times 1}$ is an input embedding, $h_{i-1} \in \real^{D_h \times 1}$ is the previous output, and  $s_{i-1} \in \real^{D_h \times 1}$ is the previous cell state.

For the left context, the LSTM is applied to the sequence $ l_{1},...,l_{C} $ from left to right and produces the outputs
$\overrightarrow{h_{1}^l},..., \overrightarrow{h_{C}^l}$.
For the right context, the sequence $ r_{C},...,r_{1} $ is processed from right to left to produce the outputs
$\overleftarrow{h_{1}^r},..., \overleftarrow{h_{C}^r}$.
The concatenation of $\overrightarrow{h_{C}^l}$ and $\overleftarrow{h_{1}^r}$ then serves as the context representation $v_c$.



\subsubsection{Attentive Encoder}

An attention mechanism aims to encourage the model to focus on salient local information that is relevant for the classification decision.
The attention mechanism variant used in this work is defined as follows.
First, bi-directional LSTMs \cite{graves2012supervised} are applied for both the right and left context.
We denote the output layers of the bi-directional LSTMs as $\overrightarrow{h_{1}^l},\overleftarrow{h_{1}^l},..., \overrightarrow{h_{C}^l},\overleftarrow{h_{C}^l}$ 
and $\overrightarrow{h_{1}^r},\overleftarrow{h_{1}^r},..., \overrightarrow{h_{C}^r},\overleftarrow{h_{C}^r}$. 

For each output layer, a scalar value $\tilde{a}_{i} \in \real$ is computed using a feed forward neural network with the hidden layer $e_i \in \real^{D_a \times 1}$ and weight matrices $W_e \in \real^{ D_a \times 2D_h }$ and $W_a \in \real^{ 1 \times D_a}$:

\begin{eqnarray}
    e_{i}^l &=& \tanh \left( W_e \left[ \begin{array}{c} \overrightarrow{h_i^l} \\ \overleftarrow{h_i^l} \\ \end{array} \right] \right) \\
    \tilde{a}_{i}^l &=& \exp(W_a e_i^l) 
\end{eqnarray}

Next, the scalar values are normalised such that they sum to $1$:

\begin{equation}
    a_{i}^l = \frac{\tilde{a}_{i}^l}{\sum_{i=1}^C \tilde{a}_{i}^l + \tilde{a}_{i}^r}
\end{equation}

These normalised scalar values $a_i \in \real$ are referred to as attentions.
Finally, we compute the sum of the output layers of the bidirectional LSTMs, weighted by the attentions $a_i$ as the representation of the context:

\begin{equation}
    v_{c} = \sum_{i=1}^{C} a_{i}^l  \Biggl[ \begin{array}{c} \overrightarrow{h_i^l} \\ \overleftarrow{h_i^l} \\ \end{array} \Biggr] 
           + a_{i}^r  \Biggl[ \begin{array}{c} \overrightarrow{h_i^r} \\ \overleftarrow{h_i^r} \\ \end{array} \Biggr]
\end{equation}

An illustration of the attentive encoder model variant can be found in Figure \ref{fig:model}.

\subsection{Hybrid Models}

To allow model variants to use both human background knowledge through hand-crafted features as well as features learnt from data, we extended the neural models to create new hybrid model variants as follows.
Let $v_f \in \real^{D_l \times 1}$ be a low-dimensional projection of the sparse feature $f(m)$:

\begin{equation}
    v_f = W_f f(m)
\end{equation}

Where $W_f \in \real^{D_l \times D_f}$ is a projection matrix.
The hybrid model variants are then defined as follows:

\begin{equation}
    y = \frac{1}{1 + \exp \left( -W_y \left[ \begin{array}{c} v_{m} \\ v_{c}\\  v_{f} \\ \end{array} \right] \right)}
\end{equation}

These models can thus draw upon learnt features through $v_m$ and $v_c$ as well as hand-crafted features using $v_f$ when making classification decisions.
While existing work on fine-grained entity type classification have used 
either sparse, manually designed features or dense, automatically learnt embedding vectors, our work is the first to propose and evaluate a model using the combination of both features.

\subsection{Hierarchical Label Encoding}

Since the fine-grained types tend to form a forest of type hierarchies (e.g. {\tt musician} is a subtype of {\tt artist}, which in turn is a subtype of {\tt person}), we investigated whether the encoding of each label could utilise this structure to enable parameter sharing.
Concretely, we compose the weight matrix $W_y$ for the logistic regression layer as the product of a learnt weight matrix $V_y$ and a constant sparse binary matrix $S$:

\begin{equation}
    W_y^{T} = V_y S
\end{equation}

\begin{figure}
    \centering
    \includegraphics[width=0.95\linewidth]{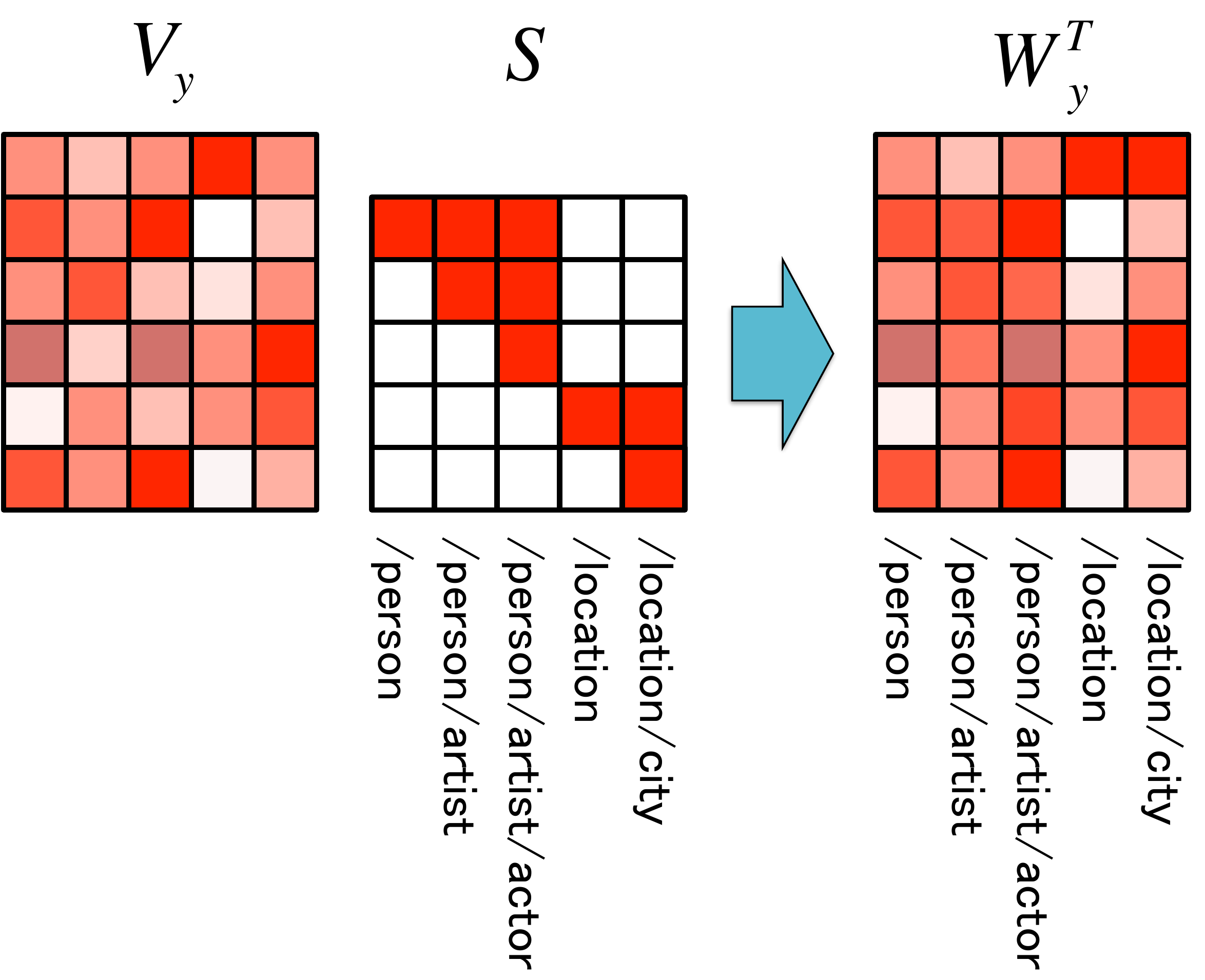}
    \caption{Hierarchical label encoding illustration.}
    \label{fig:hierarchical}
\end{figure}

We encode the type hierarchy formed by the set of types in the binary matrix $S$ as follows.
Each type is mapped to a unique column in $S$, where membership at each level of its type hierarchy is marked by a $1$.
For example, if we use the set of types defined by \newcite{gillick2014context}, the column for {\tt /person} could be encoded as $[1, 0, \ldots]$, \texttt{/person/artist} as $[1, 1, 0, \ldots]$, and \texttt{/person/artist/actor} as $[1, 1, 1, 0, \ldots]$.
This encoding scheme is illustrated in Figure~\ref{fig:hierarchical}.

This enables us to share parameters between labels in the same hierarchy, potentially making learning easier for infrequent types that can now draw upon annotations of other types in the same hierarchy.

\section{Experiments}

\subsection{Datasets}

\begin{table}
\centering
\smaller[2]
\begin{tabular}{lccccc}
    Work    & W2M   & W2M+D & W2.6M & GN1   & GN2   \\
    \toprule
    \newcite{ling2012fine} & \checkmark & &   &   &   \\
    \newcite{gillick2014context} &   &    &   & $\times$    &   \\
    \newcite{yogatama2015embedding} &   &   &   &   & $\times$  \\
    \newcite{ren2016label} & \checkmark & $\times$  &   &   &   \\
    \newcite{shimaoka2016attentive} &   &   & \checkmark &   &   \\
\end{tabular}
\caption{
    Training datasets used and its availability.
    W2M and W2.6M are Wikipedia-based, +D indicates denoising, and GN1/GN2 are two company-internal Google News datasets.
    The symbols \checkmark and $\times$ indicates publicly available and unavailable data.
}
\label{tbl:datasets}
\end{table}

Despite the research community having largely settled on using the manually annotated datasets \textsc{Figer~(GOLD)} \cite{ling2012fine} and OntoNotes \cite{gillick2014context} for evaluation, there is still a remarkable difference in the data used to train models (Table~\ref{tbl:datasets}) that are then evaluated on the same manually annotated datasets.
Also worth noting is that some data is not even publicly available, making a fair comparison between methods even more difficult.
For evaluation, in our experiments we use the two well-established manually annotated datasets \textsc{Figer~(GOLD)} and OntoNotes, where like \newcite{gillick2014context}, we discarded pronominal mentions, resulting in a total of $8,963$ mentions.
For training, we use the automatically induced publicly available datasets provided by \newcite{ren2016label}.
\newcite{ren2016label} aimed to eliminate label noise generated in the process of distant supervision and we use the ``raw'' noisy data\footnote{
    Although \newcite{ren2016label} provided both ``raw'' data and code to ``denoise'' the data, we were unable to replicate the performance benefits reported in their work after running their pipeline.
    We have contacted them regarding this as we would be interested in comparing the benefit of their denoising algorithm for each model, but at the time of writing we have not yet received a response.
} provided by them for training our models.

\subsection{Pre-trained Word Embeddings}

We use pre-trained word embeddings that were not updated during training to help the model generalise to words not appearing in the training set \cite{rocktaschel2015reasoning}.
For this purpose, we used the freely available $300$-dimensional cased word embeddings trained on 840 billion tokens from the Common Crawl supplied by \newcite{pennington2014glove}.
For words not present in the pre-trained word embeddings, we use the embedding of the ``unk'' token.

\subsection{Evaluation Criteria}

We adopt the same criteria as \newcite{ling2012fine}, that is, we evaluate the model performance by strict accuracy, loose macro, and loose micro scores.

\subsection{Hyperparameter Settings}

Values for the hyperparameters were obtained from preliminary experiments by evaluating the model performance on the development sets.
Concretely, all neural and hybrid models used the same $D_m = 300$-dimensional word embeddings, the hidden-size of the LSTM was set to $D_h = 100$, the hidden-layer size of the attention module was set to $D_a = 100$, and the size of the low-dimensional projection of the sparse features was set to $D_l=50$.
We used \texttt{Adam} \cite{kingma2014adam} as our optimisation method with a learning rate of $0.001$, a mini-batch size of $1,000$, and iterated over the training data for five epochs.
As a regularizer we used dropout \cite{hinton2012improving} with probability $0.5$ applied to the mention representation and sparse feature representation.
The context window size was set to $C=10$ and if the length of a context extends beyond the sentence length, we used a padding symbol in-place of a word.
After training, we picked the best model on the development set as our final model and report their performance on the test sets.
Our model implementation was done in Python using the TensorFlow \cite{abadi2015tensorflow} machine learning library.

\subsection{Results}

When presenting our results, it should be noted that we aim to make a clear separation between results from models trained using different datasets.

\subsubsection{\textsc{Figer (GOLD)}}

\begin{table}
\centering
\smaller
\begin{tabular}{lccc}
    Model   & Acc.  & Macro & Micro \\
    \toprule
    Hand-crafted    & 51.33 & 71.91 & 68.78 \\
    \midrule
    Averaging   & 46.36 & 71.03 & 65.31 \\
    Averaging + Hand-crafted    & 52.58 & 72.33 & 70.04 \\
    \midrule
    LSTM    & 55.60 & 75.15 & 71.73 \\
    LSTM + Hand-crafted & 57.02 & 76.98 & 73.94 \\
    \midrule
    Attentive   & 54.53 & 74.76 & 71.58 \\
    Attentive + Hand-crafted    & \bf{59.68}    & \bf{78.97}    & \bf{75.36}    \\
    \midrule
    \textsc{Figer} \cite{ling2012fine}  & 52.30 & 69.90 & 69.30 \\
    \textsc{Figer} \cite{ren2016label}  & 47.4  & 69.2  & 65.5  \\
\end{tabular}
\caption{
    Performance on \textsc{Figer (GOLD)} for models using the \textit{same} W2M training data.
}
\label{tbl:figer_same}
\end{table}

\begin{table}
\centering
\smaller[2]
\begin{tabular}{llccc}
    Model   & Data & Acc.  & Macro & Micro \\
    \toprule
    Attentive + Hand-crafted    & W2M & 59.68 & \bf{78.97}    & \bf{75.36}    \\
    Attentive \cite{shimaoka2016attentive}  & W2.6M & 58.97 & 77.96 & 74.94 \\
    \midrule
    \textsc{Figer} + PLE \cite{ren2016label}    & W2M+D & \bf{59.9} & 76.3  & 74.9  \\
    HYENA + PLE \cite{ren2016label} & W2M+D & 54.2  & 69.5  & 68.1  \\
    \midrule
    K-WASABIE \cite{yogatama2015embedding}  & GN2 & n/a   & n/a   & 72.25 \\
\end{tabular}
\caption{
    Performance on \textsc{Figer~(GOLD)} for models using \textit{different} training data.
}
\label{tbl:figer_different}
\end{table}

We first analyse the results on \textsc{Figer~(GOLD)} (Tables~\ref{tbl:figer_same} and \ref{tbl:figer_different}).
The performance of the baseline model that uses the sparse hand-crafted features is relatively close to that of the \textsc{Figer} system of \newcite{ling2012fine}. 
This is consistent with the fact that both systems use linear classifiers, similar sets of features, and training data of the same size and domain.

Looking at the results of neural models, we observe a consistent pattern that adding hand-crafted features boost performance significantly, indicating that the learnt and hand-crafted features complement each other.
The effects of adding the hierarchical label encoding were inconsistent, sometimes increasing, sometimes decreasing performance.
We thus opted not to include them in the results table due to space constraints and hypothesise that given the size of the training data, parameter sharing may not yield any large performance benefits.
Among the neural models, we see that the averaging encoder perform considerably worse than the others.
Both the LSTM and attentive encoder show strong results and the attentive encoder with hand-crafted features achieves the best performance among all the models we investigated.

When comparing our best model to previously introduced models trained using different training data, we find that we achieve state-of-the-art results both in terms of loose macro and micro scores.
The closest competitor, \textsc{Figer} + PLE \cite{ren2016label}, achieves higher accuracy at the expense of lower F1 scores, we suspect that this is due to an accuracy focus in their label pruning strategy.
It is worth noting that we achieve state-of-the-art results without the need for any noise reduction strategies.
Also, even with 600,000 fewer training examples, our variant with hand-crafted features of the attentive model from \newcite{shimaoka2016attentive} outperforms its feature-learning variant.

In regards to the impact of the choice of training set, we observe that the model introduced in \newcite{shimaoka2016attentive} drops as much as $3.36$ points of loose micro score when using a smaller dataset.
Thus casting doubts upon the comparability of results of fine-grained entity classification models using different training data.

\subsubsection{OntoNotes}

\begin{table}
\centering
\smaller[2]
\begin{tabular}{lccc}
    Model   & Acc.  & Macro & Micro \\
    \toprule
    Hand-crafted    & 48.16 & 66.33 & 60.16 \\
    \midrule
    Averaging   & 46.17 & 65.26 & 58.25 \\
    Averaging + Hier    & 47.15 & 65.53 & 58.25 \\
    Averaging + Hand-crafted    & 51.57 & 70.61 & 64.24 \\
    Averaging + Hand-crafted + Hier & \bf{51.74}    & \bf{70.98}    & 64.91 \\
    \midrule
    LSTM    & 49.20 & 66.72 & 60.52 \\
    LSTM + Hier & 48.96 & 66.51 & 60.70 \\
    LSTM + Hand-crafted & 48.58 & 68.54 & 62.89 \\
    LSTM + Hand-crafted + Hier  & 50.42 & 69.99 & 64.57 \\
    \midrule
    Attentive   & 50.32 & 67.95 & 61.65 \\
    Attentive + Hier    & 51.10 & 68.19 & 61.57 \\
    Attentive + Hand-crafted    & 49.54 & 69.04 & 63.55 \\
    Attentive + Hand-crafted + Hier & 50.89 & 70.80 & \bf{64.93}    \\
    \midrule
    \textsc{Figer} \cite{ren2016label}  & 36.90 & 57.80 & 51.60 \\
\end{tabular}
\caption{
    Performance on OntoNotes for models using the \textit{same} W2M training data.
}
\label{tbl:onto_same}
\end{table}

Secondly, we discuss the results on OntoNotes (Tables~\ref{tbl:onto_same}, and \ref{tbl:onto_different}).
Again, we see consistent performance improvements when the sparse hand-crafted features are added to the neural models. 
In the absence of hand-crafted features, the averaging encoder suffer relatively poor performance and the attentive encoder achieves the best performance.
However, when the hand-crafted features are added, a significant improvement occurs for the averaging encoder, making the performance of the three neural models much alike.
We speculate that some of the hand-crafted features such as the dependency role and parent word of the head noun, provide crucial information for the task that cannot be captured by the plain averaging model, but can be learnt if an attention mechanism is present. 
Another speculative reason is that because the training dataset is noisy  compared to \textsc{Figer~(GOLD)} (since \textsc{Figer~(GOLD)} uses anchors to detect entities whereas OntoNotes uses an external tool), and the size of the dataset is small, the robustness of the simpler averaging model becomes clearer when combined with the hand-crafted features.

Another interesting observation can be seen for models with the hierarchical label encoding, where it is clear that consistent performance increases occur.
This can be explained by the fact that the type ontology used in OntoNotes is more well-formed than its \textsc{Figer} counterpart.
While we do not obtain state-of-the-art performance when considering models using different training data, we do note that in terms of F1-score we perform within 1 point of the state of the art.
This being achieved despite having trained our models on different non-proprietary noisy data.

\begin{table}
\centering
\smaller[2]
\begin{tabular}{llccc}
    Model   & Data  & Acc.  & Macro & Micro \\
    \toprule
    Averaging + Hand-crafted + Hier & W2M   & 51.74 & 70.98 & 64.91 \\
    Attentive + Hand-crafted + Hier & W2M   & 50.89 & 70.80 & 64.93 \\
    \midrule
    \textsc{Figer} + PLE \cite{ren2016label}    & W2M+D & \bf{57.2} & \bf{71.5} & 66.1  \\
    HYENA + PLE \cite{ren2016label} & W2M+D & 54.6  & 69.2  & 62.5  \\
    \midrule
    Hand-crafted \cite{gillick2014context}  & GN1   & n/a   & n/a   & 70.01 \\
    K-WASABIE \cite{yogatama2015embedding}  & GN2   & n/a   & n/a   & \bf{72.98}    \\
\end{tabular}
\caption{
    Performance on OntoNotes for models using \textit{different} training data.
}
\label{tbl:onto_different}
\end{table}

Once again we have an opportunity to study the impact of the choice of training data by comparing the results of the hand-crafted features of \newcite{gillick2014context} to our own comparable set of features.
What we find is that the performance drop is very dramatic, $9.85$ points of loose micro score.
Given that the training data for the previously introduced model is not publicly available, we hesitate to speculate as to exactly why this drop is so dramatic, but similar observations have been made for entity linking \cite{ling2015design}.
This clearly underlines how essential it is to compare models on an equal footing using the same training data.

\subsection{PCA visualisation of label embeddings}

\begin{figure}
    \centering
    \includegraphics[width=8cm]{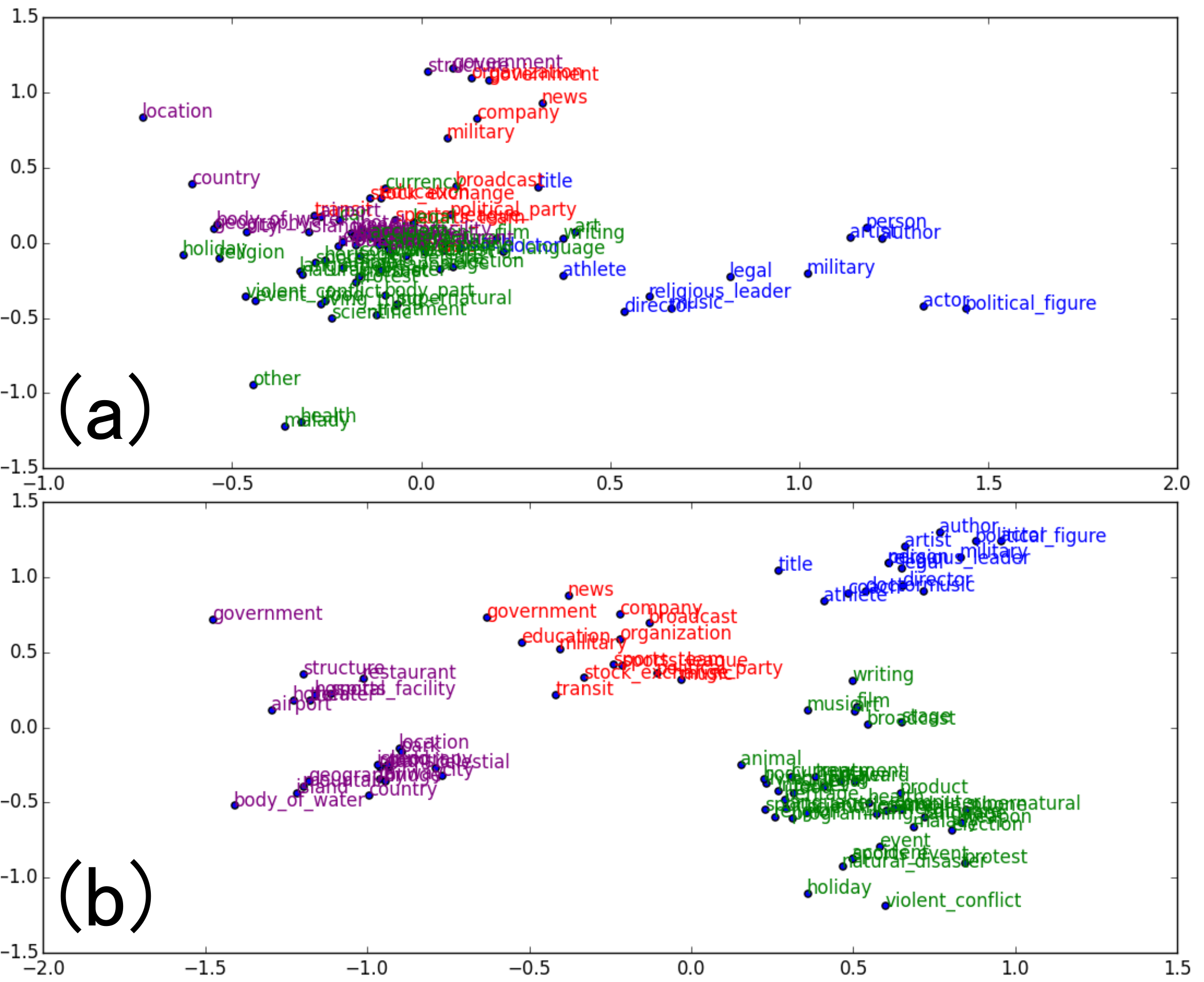}
    \caption{
        PCA projections of the label embeddings learnt from the OntoNotes dataset where subtypes share the same color as their parent type.
        Sub-figure (a) uses the non-hierarchical encoding, while sub-figure (b) uses the hierarchical encoding.
    }
    \label{fig:pca}
\end{figure}

By visualising the learnt label embeddings (Figure~\ref{fig:pca}) and comparing the non-hierarchical and hierarchical label encodings, we can observe that the hierarchical encoding forms clear distinct clusters.

\subsection{Attention Analysis}


While visualising the attention weights for specific examples have become commonplace, it is still not clear exactly what syntactic and semantic patterns that are learnt by the attention mechanism.
To better understand this, we first qualitatively analysed a large set of attention visualisations and observed that head words and the words contained in the phrase forming the mention tended to receive the highest level of attention.
In order to quantify this notion, we calculated how frequently the word strongest attended over for all mentions of a specific type was the syntactic head or the words before and after the mention in its phrase.
What we found through our analysis (Table~\ref{table:associations}) was that our attentive model without hand-crafted features does indeed learn that head words and the phrase surrounding the mention are highly indicative of the mention type, without any explicit supervision.
Furthermore, we believe that this in part might explain why the performance benefit of adding hand-crafted features was smaller for the attentive model compared to our other two neural variants.


\begin{table}
\centering
\smaller
\begin{tabular}{lllll}
    Type    & Parent    & Before    & After & Frequent Words    \\
    \toprule
    /location   & 0.319 & 0.228 & 0.070  & in, at, born  \\
    /organization   & 0.324 & 0.178 & 0.119 & at, the, by   \\
    /art/film   & 0.207 & 0.429 & 0.021 & film, films, in   \\
    /music  & 0.259 & 0.116 &   0.018   & album, song, single   \\
    /award  & 0.583 & 0.292 &   0.083   & won, a, received  \\
    /event  & 0.310 & 0.188 &   0.089   & in, during, at    \\
\end{tabular}
\caption{Quantitative attention analysis.}
\label{table:associations}   
\end{table}

\section{Conclusions and Future Work}

In this paper, we investigated several model variants for the task of fine-grained entity type classification.
The experiments clearly demonstrated that the choice of training data -- which until now been ignored for our task -- has a significant impact on performance.
Our best model achieved state-of-the-art results with 75.36\% loose micro F1 score on \textsc{Figer (GOLD)} despite being compared to models trained using larger datasets and we were able to report the best results for any model trained using publicly available data for OntoNotes with 64.93\% loose micro F1 score.
The analysis of the behaviour of the attention mechanism demonstrated that it can successfully learn to attend over expressions that are important for the classification of fine-grained types.
It is our hope that our observations can inspire further research into the limitations of what linguistic phenomena attentive models can learn and how they can be improved.

As future work, we see the re-implementation of more methods from the literature as a desirable target, so that they can be evaluated after utilising the same training data.
Additionally, we would like to explore alternative hierarchical label encodings that may lead to more consistent performance benefits.

To ease the reproducability of our work, we make our code used for the experiments available at \url{https://github.com/shimaokasonse/NFGEC}. 

\ifaclfinal
\section*{Acknowledgments}
This work was supported by CREST-JST, JSPS KAKENHI Grant Number 15H01702, a Marie Curie Career Integration Award, and an Allen Distinguished Investigator Award.
We would like to thank Dan Gillick for answering several questions related to his 2014 paper and the anonymous reviewers for their helpful feedback and encouragement.
\fi

\bibliography{figer}
\bibliographystyle{eacl2017}

\end{document}